\documentclass[runningheads]{llncs}
\usepackage{graphicx}
\usepackage{comment}
\usepackage{amsmath,amssymb} %
\usepackage{color}
\usepackage[font=small,labelfont=bf]{caption}
\usepackage{microtype}

\usepackage{graphicx}
\usepackage{subcaption}
\usepackage{float}
\usepackage{lscape}                                         %
\usepackage{wrapfig}

\usepackage[lined,ruled,linesnumbered]{algorithm2e}
\usepackage{animate} %

\usepackage{booktabs}                   %
\usepackage{multirow}

\usepackage{makecell}

\usepackage{paralist}
\usepackage{enumitem}

\usepackage{bm}                          %
\usepackage{epsfig}                      %
\usepackage{graphicx}                  %
\usepackage{mathtools}
\usepackage{amssymb,amsmath}   %

\usepackage{units}
\usepackage{color}

\usepackage{comment}

\usepackage{url}  %
\usepackage[pagebackref=true,breaklinks=true,letterpaper=true,colorlinks,bookmarks=false]{hyperref}

\usepackage{xspace}
\usepackage[table]{xcolor}
\usepackage{setspace}

\def\etal{et~al.}			  %
\def\eg{e.g.,~}               %
\def\ie{i.e.,~}               %

\newlength\paramargin
\newlength\figmargin

\newlength\secmargin
\newlength\figcapmargin
\newlength\tabcapmargin
\newlength\remarkmargin

\newcommand{\red}{\textcolor{red}}
\newcommand{\blue}{\textcolor{blue}}

\newcommand{\mpage}[2]
{
\begin{minipage}{#1\linewidth}\centering
#2
\end{minipage}
}

\newcommand{\figref}[1]{Figure~\ref{fig:#1}} 
\newcommand{\tabref}[1]{Table~\ref{tab:#1}}

\long\def\ignorethis#1{}
\newcommand {\jiabin}[1]{{\color{blue}\textbf{Jia-Bin: }#1}\normalfont}

\newcommand{\tb}[1]{\textbf{#1}}

\newcommand{\final}{0}
\ifthenelse{\equal{\final}{1}}
{
\renewcommand{\jiabin}[1]{}

}
{}

\newbox\jsavebox%

\newcommand{\best}[1]{\red{\textbf{#1}}}
\newcommand{\second}[1]{\blue{\underline{#1}}}

\newcommand{\para}[1]{
\vspace{1mm}
\noindent\textbf{#1}
}

\def\xi{\mathbf{x}_i}

\graphicspath{{figure}, {example}}

\usepackage{animate} %

\usepackage[width=122mm,left=12mm,paperwidth=146mm,height=193mm,top=12mm,paperheight=217mm]{geometry}

\begin{document}
\pagestyle{headings}
\mainmatter
\def\ECCVSubNumber{2256}  %

\title{
Learning Monocular Visual Odometry via Self-Supervised Long-Term Modeling
} %

\titlerunning{Self-Supervised Long-Term Monocular Visual Odometry}
\author{
Yuliang Zou\inst{1}
\and
Pan Ji\inst{2}
\and
Quoc-Huy Tran\inst{2}
\and
\\
Jia-Bin Huang\inst{1}
\and
Manmohan Chandraker\inst{2,3}
}
\authorrunning{Y. Zou et al.}
\institute{$^1$Virginia Tech~\quad~$^2$NEC Labs America~\quad~$^3$UCSD}
\maketitle
\begin{center}
    \centering
    \includegraphics[width=0.70\linewidth]{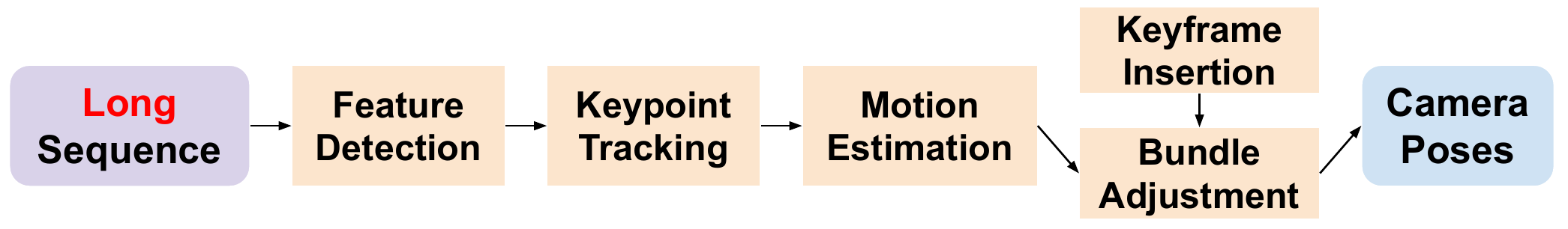}
    
    \vspace{-1.2mm}
    {\small (a) Traditional geometry-based methods}
    \\
   \mpage{0.33}{\includegraphics[width=\linewidth]{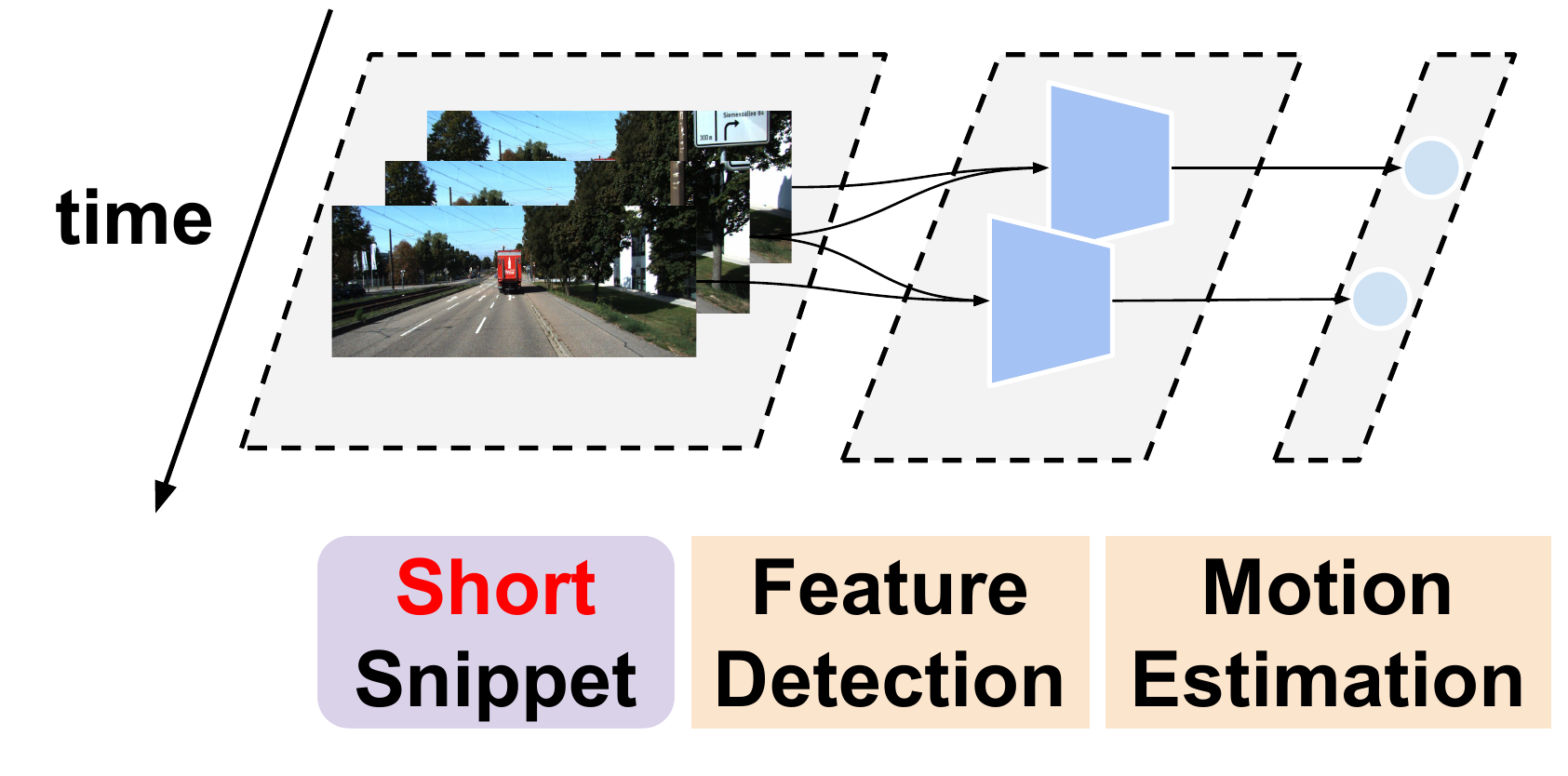}
    }\hspace{1.0cm}%
    \mpage{0.5}{\includegraphics[width=\linewidth]{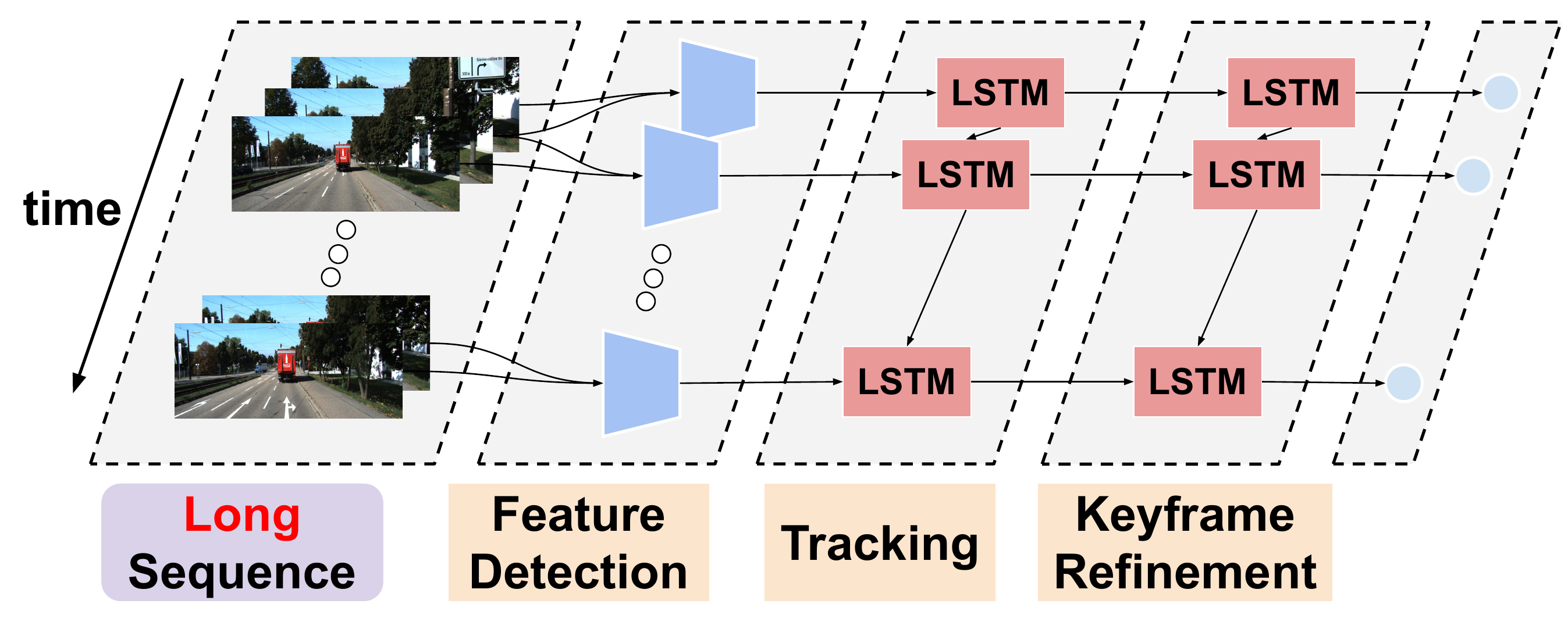}
    }\hfill
    
    \vspace{-3.0mm}
    \mpage{0.45}{{\small (b) Existing self-supervised methods}}\hspace{-0.5cm}%
    \mpage{0.5}{{\small (c) Our proposed method}}\hfill
    \vspace{-2.5mm}
    \captionof{figure}{\textbf{
    Learning monocular visual odometry with long-term modeling.
    }
    Existing self-supervised methods only see \textit{short} snippets during the \textit{training} time, which makes it hard to learn to leverage temporal consistency over \textit{long} sequences.
    Our method, in contrast, inspired by geometry-based visual odometry methods, combines the best of both the geometry and learning to aggregate long-term temporal information.
    }
    \label{fig:teaser}
\end{center}

\begin{abstract}
Monocular visual odometry (VO) suffers severely from error accumulation during frame-to-frame pose estimation.
In this paper, we present a self-supervised learning method for VO with special consideration for consistency over longer sequences.
To this end, we model the long-term dependency in pose prediction using a pose network that features a two-layer convolutional LSTM module.
We train the networks with purely self-supervised losses, including a cycle consistency loss that mimics the loop closure module in geometric VO.
Inspired by prior geometric systems, we allow the networks to see beyond a small temporal window during training, through a novel a loss that incorporates temporally distant (\eg $O(100)$) frames.
Given GPU memory constraints, we propose a stage-wise training mechanism, where the first stage operates in a local time window and the second stage refines the poses with a ``global'' loss given the first stage features. 
We demonstrate competitive results on several standard VO datasets, including KITTI and TUM RGB-D.
\footnote{Project page: \url{https://yuliang.vision/LTMVO/}}
\end{abstract}
\section{Introduction}
\label{sec:intro}
\vspace{\secmargin}

Most existing VO systems are either {\it geometric} or {\it learning-based}. 
In this paper, we argue that a truly robust VO system should combine the best of both worlds (i.e., geometry and learning). 
In particular, we propose a self-supervised method to learn monocular VO with long-term modeling, where the training scheme is directly inspired by traditional geometric methods (see Figure~\ref{fig:teaser}).

At the heart of the state-of-the-art VO systems~\cite{engel2017direct,engel2014lsd,forster2014svo,murORB2} is the incorporation of several long-studied geometric modules, including keypoint tracking, motion estimation, keyframe insertion, and bundle adjustment (BA)~\cite{triggs1999bundle}.
With all these modules, a key insight is to optimize the states (\eg 6-DoF camera poses) over {\it long-term} observations such that the system suffers less from error accumulation~\cite{scaramuzza2011visual}.
While being robust in normal scenarios, monocular VO still suffers from the difficulty in initialization for slow motions~\cite{mur2015orb}, and the tracking tends to fail miserably in unconstrained environments with large texture-less regions, fast movements, or other adverse factors~\cite{yang2018challenges} such as rolling shutter effect~\cite{schubert2018direct,zhuang2019learning} and unknown camera intrinsics~\cite{bogdan2018deepcalib,zhuang2019degeneracy}.

In contrast, learning-based VO methods~\cite{wang2017deepvo,wang2018end,xue2018guided,xue2019beyond} have the potential of being more robust to the aforementioned challenges by harnessing the rich priors from data. 
However, training neural networks in a supervised way involves collecting large-scale, diverse datasets with ground truth annotations, which could be labor-extensive and time-consuming.
Recently, self-supervised methods~\cite{godard2019digging,li2018undeepvo,ranjan2019competitive,wang2019recurrent,zhan2018unsupervised,zhou2017unsupervised,zou2018dfnet} have been proposed to tackle this task. 
Instead of supervising the networks with ground truth labels, the idea is to couple the depth and pose networks with photometric errors across adjacent frames and jointly train them in an end-to-end manner. 
Nonetheless, the performance of these methods still falls behind that of geometric methods~\cite{mur2015orb} for general scenarios.

\begin{figure}[t]
\centering
\mpage{0.28}{\includegraphics[width=\linewidth]{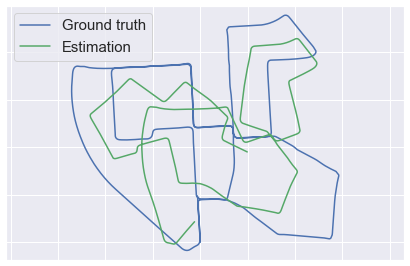}}\hfill
\mpage{0.28}{\includegraphics[width=\linewidth]{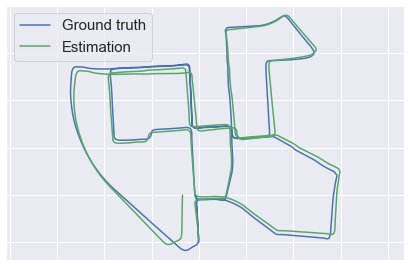}}\hfill
\mpage{0.4}
{
\resizebox{1\textwidth}{!}{
\begin{tabular}{lcc}
\toprule
Evaluation & 5-frame & video-level \\
\midrule
SfMLearner \cite{zhou2017unsupervised} & 0.019 & 97.81 \\

Our method & 0.015 & 13.13 \\
\bottomrule
\end{tabular}
}
}\hfill

\vspace{-2.0mm}
\mpage{0.3}{\small (a) 
SfMLearner~\cite{zhou2017unsupervised}
} \hfill
\mpage{0.2}{\small (b) Our method} \hfill
\mpage{0.4}{\small (c) RMSE (m)} \hfill
\vspace{-2mm}
\caption{\textbf{5-frame v.s. video-level evaluation.} 
Evaluating visual odometry requires having a global picture of recovered trajectories.
However, most self-supervised methods only evaluate the trajectories within a short snippet, which may not reflect the holistic performance.
``5-frame" means that we evaluate the results using 5-frame snippets, and ``video-level" means that we evaluate on the entire trajectory.
}
\label{fig:motivation}
\end{figure}

One of the potential reasons for their performance gap is that the pose networks do not exploit the temporal coherence over long sequences.
During training, these networks receive \textit{short} snippets (e.g., 3-frame or 5-frame) as input and predict the ego-motions that are optimized \textit{locally} for the current snippet.
When evaluating these methods in short snippets, they compare favorably even with the state-of-the-art geometric methods~\cite{mur2015orb}.
However, if we concatenate all the predictions to form the full trajectory, it is often the case that the learning-based methods generate much larger pose errors, as illustrated in Figure~\ref{fig:motivation}.

In this paper, we argue that learning VO requires explicit \textit{long-term} modeling to infuse the insights from geometric methods~\cite{engel2017direct,engel2014lsd,forster2014svo,murORB2}.
To this end, we propose a novel self-supervised VO learning framework that draws inspiration from geometric modules. 
Specifically, we build our learning framework upon a depth network of an auto-encoder structure with skip-connections~\cite{godard2019digging} and a pose network with a two-layer LSTM module~\cite{xue2019beyond}. 
In contrast to the supervised method by Xue~\etal~\cite{xue2019beyond}, our method incorporates extra depth information and uses a completely different training scheme, leading to a purely self-supervised learning framework.
To mitigate error accumulation, we propose a cycle consistency constraint between the two-layer predictions, mimicking a mini {\it loop closure} module, which improves the pose consistency over the sequence.
In order to model long-term dependency in VO, we propose a two-stage training strategy, which considers both short-term and long-term constraints.
The proposed two training stages correspond to the {\it local} and {\it global} bundle adjustment modules in the geometric VO, allowing us to refine the poses within a large temporal range.

In summary, our contributions are:
\begin{itemize}[noitemsep]
\item We propose a novel self-supervised VO learning framework that explicitly models long-term temporal dependency.

\item We build connections between our method and key building blocks of geometric VO systems and demonstrate well-motivated designs.

\item We evaluate the {\it full} pose trajectories by our method, against the state-of-the-art geometric and learning-based baselines, and achieve competitive results on standard VO datasets, including KITTI and TUM RGB-D.
\end{itemize}

To the best of our knowledge, our method is the first of the kind that is able to learn from ``truly'' long sequences (\eg $\sim$100 frames) in the training stage.
Our experiments show that our proposed method gives rise to significant empirical benefits by explicitly considering long-term modeling.

\section{Related Work}
\label{sec:related}
\vspace{\secmargin}

\para{Geometric Methods.}
Visual odometry is a long-standing problem that estimates the ego-motion incrementally~\cite{nister2004visual,scaramuzza2011visual} using visual input.
A conventional geometric VO system usually consists of the following components~\cite{scaramuzza2011visual}: feature detection, feature matching (or tracking), motion estimation (\eg triangulation~\cite{hartley1997triangulation}), and local optimization (\eg bundle adjustment). 
A keyframe mechanism~\cite{klein2007parallel} is also adopted for improved robustness in motion estimation.
Incorporated with a mapping system that reconstructs the 3D scene structures, a VO system turns to a system called Simultaneous Localization and Mapping (SLAM)~\cite{cadena2016past}. 
The key to the robustness of the modern VO/SLAM systems~\cite{mur2015orb,tiwari2020pseudo} lies in their capability to extract reliable image measurements and optimize the states (\eg 6-DoF camera poses) over a large number of frames. 
In this work, we leverage these geometric insights to design a robust learning-based VO system.

\para{Fully-Supervised Methods.}
With the success of deep neural networks, end-to-end learning-based methods~\cite{wang2017deepvo,wang2018end,xue2018guided,xue2019beyond} have been proposed to tackle the visual odometry problem. 
These methods often rely on a supervised loss using the ground-truths to regress the 6-DoF camera relative pose from a pair of consecutive images.
Recently, some methods~\cite{bloesch2018codeslam,tang2018ba,teed2018deepv2d,ummenhofer2017demon,zhou2018deeptam} exploit CNNs to predict the scene depth and camera pose jointly, utilizing the geometric connection between the structure and the motion. 
This corresponds to learning Structure-from-Motion (SfM) in a supervised manner.
Although the methods above achieve good performance, they require ground-truth annotations to train the networks.
In contrast, our method is self-supervised, requiring nothing but the monocular video frames.

\para{Self-Supervised Methods.}
To mitigate the requirement of data annotations, self-supervised methods~\cite{godard2019digging,li2018undeepvo,ranjan2019competitive,wang2019recurrent,zhan2018unsupervised,zhou2017unsupervised} have been proposed to tackle the SfM task. 
The main supervisory signal of these methods comes from the photometric-consistency between corresponding pixels of neighboring frames.
While they achieve good performance on single-view depth estimation, the performance of ego-motion estimation still lags behind the traditional SLAM/VO methods.
Recently, Bian~\etal~\cite{bian2019unsupervised} argue that the pose networks cannot provide full camera trajectories over long sequences due to the inconsistent scale of per-frame estimations and thus propose a geometry consistency constraint.
However, their method only enforces the globally consistent trajectories by propagating the consistency on overlapping \textit{short} snippets during training.
In contrast, our method directly optimizes over \textit{long} sequences via long-term modeling.
Inspired by the keyframe mechanism in geometric methods, Sheng~\etal~\cite{sheng2019unsupervised} propose to jointly learn depth, ego-motion, and keyframe selection simultaneously in a self-supervised manner.
Similarly, the training of this method considers only \textit{short} snippets and thus is unable to model long-term dependency.

\para{Sequential Modeling.}
Sequential modeling based on recurrent neural networks (RNNs) has been successfully applied to many applications, such as speech recognition~\cite{chorowski2014end}, machine translation~\cite{graves2014towards}, and video prediction~\cite{srivastava2015unsupervised,villegas2017learning}.
Aiming to estimate the full trajectory over a long sequence of frames, VO can be naturally formulated as a sequential learning problem and thus modeled with RNNs~\cite{wang2019recurrent,wang2017deepvo,wang2018end,xue2018guided}.
Recently, Xue~\etal~\cite{xue2019beyond} propose to use a two-layer LSTM network for pose estimation, where the first layer estimates the relative motion between consecutive frames, and the second layer estimates global absolute poses.

Despite using a similar pose network, our method differs from Xue~\etal~\cite{xue2019beyond} in being self-supervised v.s. full-supervised and the associated training strategies.
First of all, our method further incorporates depth information, while the method in \cite{xue2019beyond} relies only on pose features.
Apart from the photometric discrepancy as to the supervisory signal, we enforce a cycle consistency between the predictions from the two-layer LSTM modules, which serves as a mini ``loop closure" module that mimics the geometry VO system.
More importantly, we decouple our network training into two stages, allowing our method to optimize over \textit{long} sequences (more than 90 frames) during training, whereas the method in~\cite{xue2019beyond} only trains with 11-frame snippets.
To our knowledge, this is the {\it first} deep learning approach for visual odometry that takes \textit{long} sequences as input in the training stage.

\section{Method}
\label{sec:method}
\vspace{\secmargin}
\begin{figure*}[t]
\mpage{0.49}{\includegraphics[width=\linewidth]{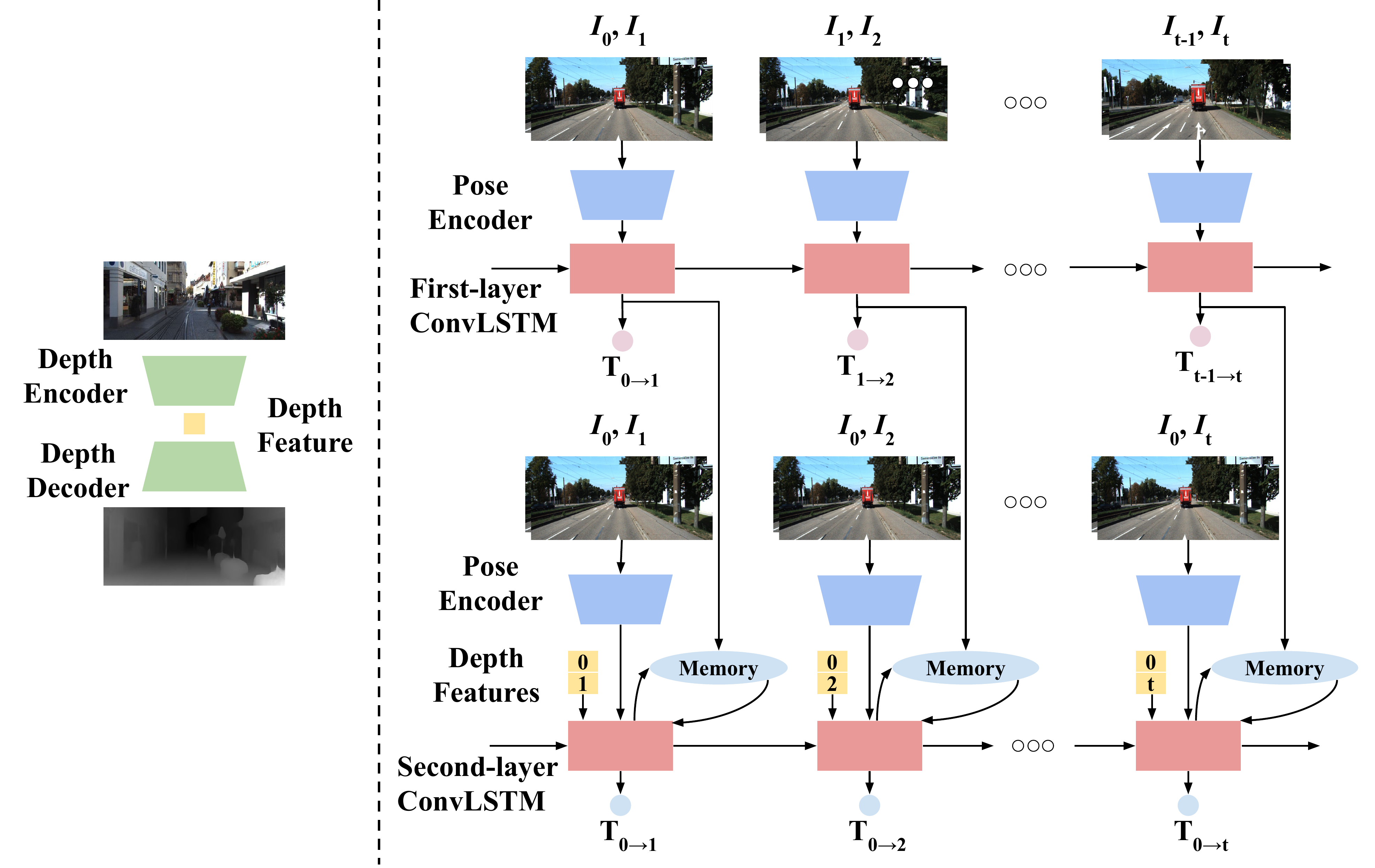}}\hfill
\mpage{0.49}{\includegraphics[width=\linewidth]{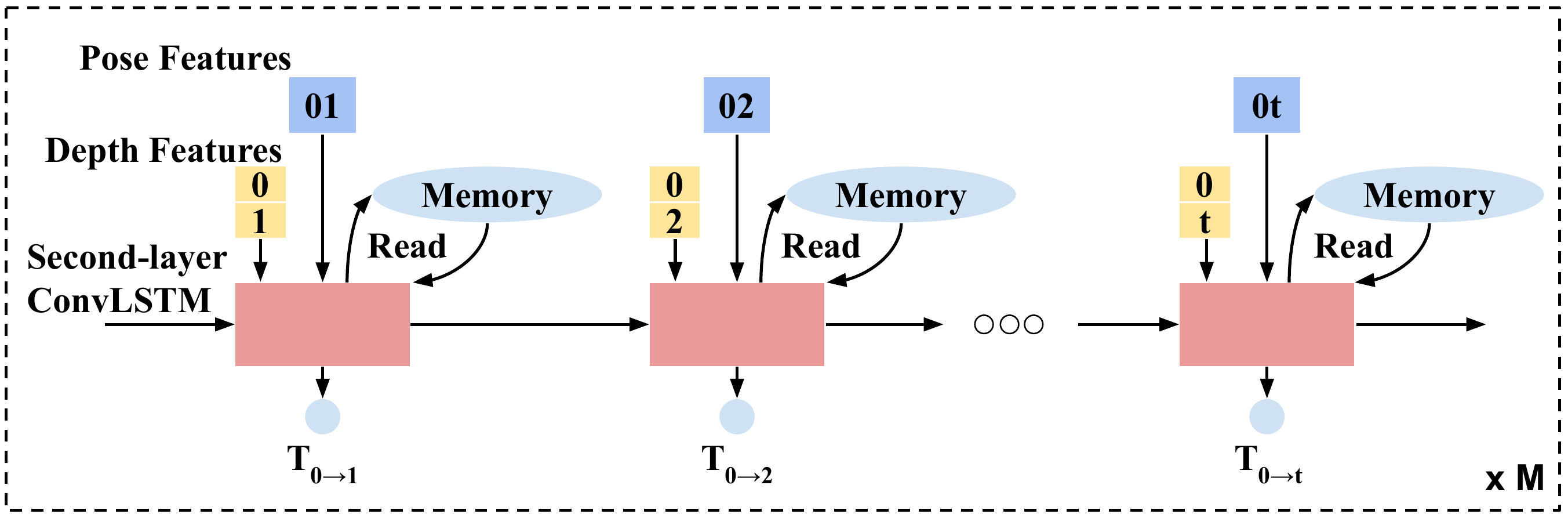}}\hfill

\mpage{0.49}{\small (a) 1$^{\rm st}$ stage of training (\textit{short} snippet)} \hfill
\mpage{0.49}{\small (b) 2$^{\rm nd}$ stage of training (\textit{long} sequence)} \hfill

\vspace{-5.5mm}
\caption{\textbf{Overview.} 
Our method adopts a stage-wise training strategy. 
(a) In the first stage of training, we jointly train all the components: the depth encoder, the depth decoder, the pose encoder, the first and second layer of ConvLSTM (Sec. \red{3.2}); 
(b) In the second stage of training, we pre-extract features as input and fine-tune the second layer of the ConvLSTM module only (Sec. \red{3.3}).
}
\label{fig:overview}
\end{figure*}

\figref{overview} provides a high-level overview of the proposed monocular VO system.
Our system has two major components: a depth network and a pose network.\footnote{Since accurate pose prediction is the primary focus of this paper, we name our method as VO instead of SfM or SLAM.}
The single-image depth network employs an encoder-decoder structure with skip-connections~\cite{godard2019digging}.
The pose network consists of a FlowNet backbone~\cite{dosovitskiy2015flownet}, a two-layer LSTM module~\cite{xue2019beyond}, and two pose prediction heads (with one after each of the LSTM layers). 
In the two-layer recurrent architecture, the first-layer focuses on predicting consecutive frame motions, while the second-layer refines the estimations from the first-layer~\cite{xue2019beyond}. 

\subsection{Background}
We formulate the monocular visual odometry task as a view synthesis problem, by training the networks to predict a target image from the source image with the estimated depth and camera pose. 
Such a system typically consists of two components: a depth network which takes a single RGB image as input to predict the depth map, and a pose network which takes a concatenation of two consecutive frames as input to estimate the 6-DoF ego-motion.

Given two input images $I_t$ and $I_{t+1}$, the estimated depth map $\hat{D}_t$ and camera pose $\hat{T}_{t\rightarrow t+1}$, we can then compute the per-pixel correspondence between the two input images.
Assume a known camera intrinsic matrix $K$, and let $p_t$ represent the 2D homogeneous coordinate of a pixel in $I_t$.
We can find the corresponding point of $p_t$ in $I_{t+1}$ 
following the equation~\cite{zhou2017unsupervised}:%
\begin{equation}
    \label{eq:warping}
    p_{t+1} \sim K \hat{T}_{t\rightarrow t+1} \hat{D}_t(p_t) K^{-1} p_t\;.
\end{equation}

\para{Appearance loss.}
For a self-supervised visual odometry system, the primary supervision comes from the appearance dissimilarity between the synthesis image and the target image.
To effectively handle occlusion, we use three consecutive frames to compute the per-pixel minimum photometric reprojection loss~\cite{godard2019digging}, \ie
\begin{equation}
    \label{eq:photometric-loss}
    L_A = \frac{1}{N-2} \sum_{t=1}^{N-2} \text{min}_{t'\in\{t-1, t+1\}} \rho(I_t, \hat{I}_{t'\rightarrow t})\;,
\end{equation}
where $\rho$ is a weighted combination of the L2 loss and the Structured SIMilarity (SSIM) loss, $\hat{I}_{t'\rightarrow t}$ denotes the frame synthesized from $I_{t'}$
using Eq.~\eqref{eq:warping}.
To handle static pixels, we adopt the auto-masking mechanism following Godard~\etal~\cite{godard2019digging}.

\para{Smoothness loss.}
Since the appearance loss cannot provide meaningful supervision for texture-less or homogeneous regions of the scene, a smoothness prior of disparity is incorporated.
We here use the edge-aware smoothness loss
$L_S$ as in Wang~\etal~\cite{wang2018learning}.

\vspace{\remarkmargin}
\begin{remark}
{\it  The appearance loss in Eq.~\eqref{eq:photometric-loss} corresponds to a local photometric bundle adjustment objective, which is also commonly used in the geometric direct VO/SLAM systems~\cite{engel2017direct,engel2014lsd,wang2017stereo}.}
\end{remark}

\subsection{Cycle consistency within memory-aided sequential modeling }
With the above setting, current state-of-the-art self-supervised methods estimate the ego-motion within a \textit{local} range, discarding the sequential dependence and dynamics in the \textit{long} sequences.
Such information, however, is essential for a pose network to recover the entire trajectory in a consistent manner.
We thus adopt a recurrent structure of our pose network to utilize the temporal information.

\para{Sequential modeling.}
To learn to utilize the temporal information, we adopt the recurrent network structure with a convolution LSTM (ConvLSTM) module~\cite{xingjian2015convolutional}.
Previously, the pose network takes the concatenation of two frames and outputs the 6-DoF camera pose directly.
After incorporating the ConvLSTM module, the pose network also takes the previous estimation information into account when predicting the output.
Formally, we have
\begin{align}
    &F_t = \text{PEnc}(I_t, I_{t-1})\;, \\
    &O_t, H_t = \text{ConvLSTM}(F_t, H_{t-1})\;, \\
    &\hat{T}_{t-1\rightarrow t} = g_1(O_t)\;,
\end{align}
where $\text{PEnc}(\cdot)$ is the pose encoder, $O_t, H_t$ denotes the output and hidden state of ConvLSTM at time $t$, $g_1(\cdot)$ is a linear layer to predict the 6-DoF motion $\hat{T}_{t-1\rightarrow t}$.
By doing this, the network implicitly learns to aggregate temporal information and learns the motion pattern.

\para{Memory buffer and refinement.}
In the sequential modeling setting above, the pose network estimates the relative pose for every two consecutive frames.
However, the motions between consecutive frames are often tiny, which results in difficulties in extracting good features for relative pose estimation.
Thus, predicting the camera pose from a non-adjacent ``anchor'' frame to the current frame could be a better option.
Note that many traditional SLAM systems~\cite{mur2015orb,murORB2} adopt a keyframe mechanism and always compute camera poses from the current frame to the most recent keyframe.

Inspired by the keyframe mechanism, we incorporate the second-layer ConvLSTM and adopt the memory module proposed by Xue~\etal~\cite{xue2019beyond}.
After each step in the first-layer ConvLSTM, we store the hidden state tensor in a memory buffer, whose length is set to the length of the input snippet.
When we read out from the memory buffer, we compute the weighted average of all the memory slots in the memory buffer as in~\cite{xue2019beyond}.

We also compute the depth and pose features for the first frame and the current frame as additional input to the second-layer ConvLSTM.
This can be formally written as
\begin{align}
    &E_{t} = \text{DEnc}(I_t)\;, \\
    &F_{t, abs} = \text{PEnc}(I_0, I_t)\;, \\
    &O_{t, abs}, H_{t, abs} = \text{ConvLSTM}(F_{t, abs}, E_0, E_t, M_t, H_{t-1, abs})\;, \\
    &\hat{T}_{0\rightarrow t} = g_2(O_{t, abs})\;,
\end{align}
where $\text{DEnc}(\cdot)$ is the depth encoder, $M_t$ is the read-out memory, $O_{t, abs}$, $H_{t, abs}$ denote the output and hidden state from the second-layer at time $t$, and $g_2(\cdot)$ is another linear layer predicting the absolute pose in the current snippet.

\vspace{\remarkmargin}
\begin{remark}
{\it The ConvLSTMs explicitly model the sequential nature of the VO problem and meanwhile facilitate the implementation of a keyframe mechanism. Compared to the memory module by Xue~\etal~\cite{xue2019beyond}, which only considers pose features, our memory module accommodates both depth and pose features. As verified in our experiments~
\footnote{\tabref{ablation_component} in the supplementary material.},
incorporating the additional depth information in memory improves the overall performance.}
\end{remark}

\begin{figure}[t]
\centering
\animategraphics[autoplay,loop,width=\textwidth]{1}{gif_files/cycle/}{000}{003}
\vspace{-3.5mm}
\caption{
\tb{Cycle-consistency over two-layer poses.}
In our model, the first layer ConvLSTM estimates the relative pose between consecutive frames, and the second layer ConvLSTM predicts the ``absolute" pose within the current snippet.
By exploiting the transitivity of camera poses, we incorporate a cycle-consistency constraint between the two layers.
\red{Animation can be viewed in Adobe Reader.}
}
\label{fig:cycle}
\end{figure}

\para{Cycle consistency over two-layer poses.}
To train the second-layer ConvLSTM, we utilize the photometric error between the first frame and the other frames of the input snippet, \ie
\begin{equation}
    L_{A, abs} = \frac{1}{N-1} \sum_{t=1}^{N-1} \rho(I_0, \hat{I}_{t\rightarrow0})\;,
\end{equation}
where $N$ is the number of frames for the input snippet, which is set to 7 in our model.

Also, according to the transitivity of the camera transformation, we have an additional constraint to ensure the consistency between the first and second layer ConvLSTM (as shown in~\figref{cycle}), \ie
\begin{equation}
\label{eq:cycle-consistency}
    L_P = \frac{1}{N-1} \sum_{t=1}^{N-1} || \hat{T}_{0\rightarrow t} - \hat{T}_{t-1\rightarrow t} \hat{T}_{0 \rightarrow t-1} ||_2^2\;.
\end{equation}

Thus, the overall objective is
\begin{equation}
    \label{eq:full-loss}
    L_\text{full} = L_A + \lambda_1 L_S + L_{A, abs} + \lambda_2 L_P\;,
\end{equation}
where $\lambda_1, \lambda_2$ are the hyper-parameters to balance the scale of different terms, which are empirically set to $0.001$.

\vspace{\remarkmargin}
\begin{remark}
 {\it The loss in Eq.~\eqref{eq:cycle-consistency} can be thought of as a mini ``loop closure'' module that enforces the cycle-consistency between the outputs of two ConvLSTM layers. Note that our method is also compatible with the existing full loop closure techniques~\cite{kummerle2011g}, which we will consider in the future work.}
\end{remark}

\subsection{Long-range constraints via stage-wise training}
Although we adopt a recurrent network structure to aggregate temporal information for better performance, the network has never seen \textit{long} sequences but only \textit{short} snippets during the training time.
Thus, the network may not learn how to fully utilize the long-term temporal context.
The hurdle that prevents us from taking long sequences as input is the limited memory volume of modern GPUs.
To tackle this problem for training a long-term model, we propose a two-stage training strategy.
We first train our whole model with the full objective $L_\text{full}$ using short snippets.

Once the first stage of training is finished, we run this model on each sequence in the dataset separately, to extract the required input for the second-layer ConvLSTM and store them.
After that, we only fine-tune the lightweight second-layer ConvLSTM, without the heavy feature extraction and depth networks, which saves us a lot of memories.
By doing this, we can now feed long sequences into the network during the training time, allowing the network to better learn how to utilize the temporal context.
Since only the second-layer ConvLSTM is optimized, our loss for the second stage of training is %
\begin{equation}
    \label{eq:long-loss}
    L_{long} = \frac{1}{M} \sum_{m=0}^{M-1} \frac{1}{N-1} \sum_{t=m(N-1)+1}^{m(N-1)+N-1} \rho(I_{m(N-1)}, \hat{I}_{t\rightarrow m(N-1)})\;,
\end{equation}
where $N$ is the number of frames of each snippet, which is set to 7; $M$ is the number of snippets in the input sequence, which is set to 16.
Note that consecutive snippets have one frame in common, and thus the total number of frames in the input sequence is 97. 
The synthesized image $\hat{I}_{t\rightarrow m(N-1)}$ is a function of depth and pose, where pose encodes long-range constraints through hidden states of ConvLSTMs, yielding an effective window of 97 frames.
We summarize our method in Algorithm~\ref{alg:stage-wise-training}.

\vspace{\remarkmargin}
\begin{remark}
\textit{
The second training stage can be viewed as a motion-only bundle adjustment module~\cite{mur2015orb} that considers long-term modeling.
}
\end{remark}

\begin{algorithm}[!tb]
    \SetKwInOut{Input}{Input}
    \SetKwInOut{Trainable}{Trainable}
    \SetKwInOut{Objective}{Objective}

    First stage: short-range training \;
    \Input{7-frame snippet}
    \Trainable{Depth Encoder, Depth Decoder, Pose Encoder, First-layer ConvLSTM, Second-layer ConvLSTM}
    \Objective{$L_\text{full}$ (Eq.~\eqref{eq:full-loss})}

    Second stage: long-range training \;
    \Input{97-frame sequence}
    \Trainable{Second-layer ConvLSTM}
    \Objective{$L_\text{long}$ (Eq.~\eqref{eq:long-loss})}
    
    \caption{Stage-wise training strategy}
    \label{alg:stage-wise-training}
\end{algorithm}

\section{Experimental Results}
\label{sec:results}
\vspace{\secmargin}

\subsection{Settings}
\para{Datasets.}
To evaluate our method, we conduct the main experiments on the {KITTI dataset}~\cite{Geiger2013IJRR,Geiger2012CVPR}, which consists of urban and highway driving sequences for road scene understanding~\cite{song2014robust,dhiman2016continuous}.
The odometry split of KITTI is a widely used benchmark for odometry/SLAM evaluation.
It contains 22 sequences, among which Sequence 00-10 have ground truth trajectory labels, and the annotations of the remaining sequences are not publicly available.
Following Zhou~\etal\cite{zhou2017unsupervised}, we use Sequence 00-08 as our training set and validate the models on Sequence 09 and  10.
Besides, we select 18 more sequences from KITTI raw data, which have no overlaps with the odometry split, for further evaluation.
Since the ground truth trajectories of Sequence 11-21 are not available, we run ORB-SLAM2 (stereo version) to get predictions as (pseudo) ground truth for evaluation.
In addition to these outdoor scenes, we also train and evaluate our model on the {TUM RGB-D dataset}~\cite{sturm2012benchmark}.
This dataset is collected by hand-held cameras in indoor environments with challenging conditions.
We use the same train/test split as in Xue~\etal~\cite{xue2019beyond}.

\para{Evaluation metrics.}
For the KITTI dataset, we adopt the absolute trajectory RMSE and relative translation/rotation errors for all possible subsequences of length (100, 200, ..., 800 meters).
For the TUM RGB-D dataset, we use the translational RMSE as our evaluation metrics.
For self-supervised monocular methods, since the absolute scale is unknown, we align the trajectory globally using the evo toolbox~\cite{grupp2017evo}.

\para{Implementation details.}
We use ImageNet pretrained ResNet-18 as our depth encoder.
Our depth decoder structure is the same as Godard~\etal~\cite{godard2019digging}.
For the pose encoder, we take the encoder of FlowNet-S structure until the last layer, which is pre-trained on FlyingChairs~\cite{dosovitskiy2015flownet} for optical flow estimation.
We implement our system using the publicly available PyTorch framework and conduct all our experiments with a single TitanXP GPU.
For both stages, we train the network with the Adam optimizer~\cite{kingma2014adam} for 20 epochs.
The learning rate is set to 5e-5 for the first 15 epochs and drops to 5e-6 for the remaining epochs.
The input size is 640$\times$192, and the batch size is set to 2.
In the first stage of training, the number of frames is set to 7, while the number of frames is set to 97 in the second stage.
Note that the long-term optimization only happens in the training time.
At the test time, our model runs at 14.3 frames per second.

\subsection{Results}%
\begin{table*}[t]
\centering
\caption{\tb{Ablation study.}
We evaluate different variants of the proposed method on sequences 09 and 10 of the KITTI Odometry dataset~\cite{Geiger2012CVPR}.
The best performance is in \best{bold} and the second best is \second{underlined}.
}

\resizebox{1\textwidth}{!}{

\begin{tabular}{lcccccccccccc}
\toprule
&& \multicolumn{3}{c}{Seq. 09} && 
\multicolumn{3}{c}{Seq. 10}
\\
\cmidrule{3-5} \cmidrule{7-9}
Method && RMSE (m) & Rel. trans. (\%) & Rel. rot. (deg/m) && RMSE (m) & Rel. trans. (\%) & Rel. rot. (deg/m) \\

\midrule
Baseline
&& 22.71 & 7.55 & 0.028 && 17.87 & 10.43 & 0.046 \\

One-layer ConvLSTM
&& 23.45 & 5.59 & 0.016 && \second{11.93} & 7.23 & \second{0.023} \\

Two-layer ConvLSTM
&& \best{9.77} & \second{4.23} & \second{0.013} && 12.68 & \second{6.02} & \second{0.023} \\

Two-layer ConvLSTM + Two-stage training
&& \second{11.30} & \best{3.49} & \best{0.010} && \best{11.80} & \best{5.81} & \best{0.018} \\

\bottomrule
\end{tabular}

}

\label{tab:kitti_ablation}
\end{table*}

\begin{figure}[t]
\centering
\includegraphics[width=1.0\linewidth]{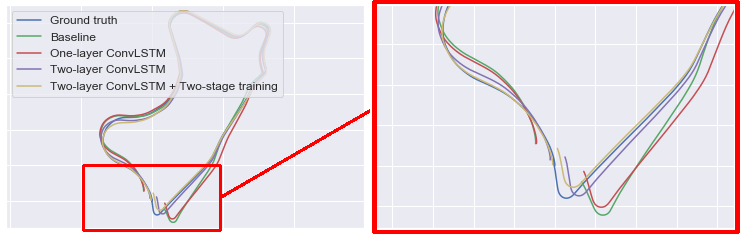}
\\
\hfill
\mpage{0.45}{\small (a) Full trajectories}\hfill
\mpage{0.45}{\small (b) Zoom-in}\hfill

\vspace{-2mm}
\caption{\textbf{Visual comparison on Sequence 09.}
We compare different variants of our method to validate the design choice.
As we can see, adding each of the components gradually improves the overall performance. Best viewed in color.
}
\label{fig:ablation}
\end{figure}
\para{Ablation study.}
To validate our design choice, we perform an ablation study on Sequence 09 and 10 of the KITTI Odometry dataset.
We consider the following variants.
1) Baseline: the pose network takes as input the concatenation of two consecutive frames to generate pose estimation;
2) One-layer ConvLSTM: we incorporate a one-layer ConvLSTM for the pose network;
3) Two-layer ConvLSTM: we use the full model, but only conduct the first stage of training;
4) Two-layer ConvLSTM + Two-stage training: our final model with the two-stage training strategy.

As shown in~\tabref{kitti_ablation}, the performance gradually improves as we add more components.
Specifically, adding a recurrent module improves the overall performance over the baseline;
adding the second layer LSTM leads to further improvement, which validates the effectiveness of the second layer in the self-supervised learning setting;
applying our second stage long-term training again boosts the performance, achieving a new state-of-the-art for self-supervised methods.
We show a visual comparison in~\figref{ablation}.

\begin{table*}[htbp]
\centering
\caption{\tb{Comparison with the state-of-the-art.}
The results of ORB-SLAM2-M methods are the medians of 5 runs.
`-' means the results are not available from that paper.
For DeepV2D~\cite{teed2018deepv2d}, SfMLearner~\cite{zhou2017unsupervised}, GeoNet~\cite{yin2018geonet}, CC~\cite{ranjan2019competitive}, DeepMatchVO~\cite{shen2019icra}, and MonoDepth2~\cite{godard2019digging}, we take the pre-trained models and run on the sequences to get the results.
The best performance of each block is in \best{bold}, and the second best is \second{underlined}.
}

\resizebox{1\textwidth}{!}{

\begin{tabular}{llccccccccccc}
\toprule
&&& \multicolumn{3}{c}{Seq. 09} && 
\multicolumn{3}{c}{Seq. 10}
\\
\cmidrule{4-6} \cmidrule{8-10}
\multicolumn{2}{c}{Method} && RMSE (m) & Rel. trans. (\%) & Rel. rot. (deg/m) && RMSE (m) & Rel. trans. (\%) & Rel. rot. (deg/m) \\

\midrule
\multirow{2}{*}{
Geo.
}
&ORB-SLAM2-M (w/o LC)
~\cite{murORB2} 
&& \second{44.10} & \second{9.67} & \best{0.003} && \best{6.43} & \best{4.04} & \best{0.003} \\

&ORB-SLAM2-M
~\cite{murORB2} 
&& \best{8.84} & \best{3.22} & \second{0.004} && \second{8.51} & \second{4.25} & \best{0.003} \\

\midrule
\multirow{6}{*}{
Sup.
}
&DeepVO
~\cite{wang2017deepvo} 
&& - & - & - && - & 8.11 & 0.088\\

&ESP-VO
~\cite{wang2018end} 
&& - & - & - && - & 9.77 & 0.102 \\

&GFS-VO
~\cite{xue2018guided} 
&& - & - & - && - & \second{6.32} & \second{0.023} \\

&GFS-VO-RNN
~\cite{xue2018guided} 
&& - & - & - && - & 7.44 & 0.032 \\

&BeyondTracking
~\cite{xue2019beyond} 
&& - & - & - && - & \best{3.94} & \best{0.017} \\

&DeepV2D
~\cite{teed2018deepv2d}
&& \best{79.06} & \best{8.71} & \best{0.037} && \best{48.49} & 12.81 & 0.083 \\

\midrule
\multirow{12}{*}{
Self-Sup.
}
&SfMLearner
~\cite{zhou2017unsupervised} 
&& 24.31 & 8.28 & 0.031 && 20.87 & 12.20 & \second{0.030} \\

&GeoNet
~\cite{yin2018geonet} 
&& 158.45 & 28.72 & 0.098 && 43.04 & 23.90 & 0.090 \\

&Depth-VO-Feat
~\cite{zhan2018unsupervised} 
&& - & 11.93 & 0.039 && - & 12.45 & 0.035 \\

&vid2depth
~\cite{mahjourian2018unsupervised} 
&& - & - & - && - & 21.54 & 0.125 \\

&UnDeepVO
~\cite{li2018undeepvo} 
&& - & 7.01 & 0.036 && - & 10.63 & 0.046 \\

&Wang~\etal
~\cite{wang2019recurrent} 
&& - & 9.88 & 0.034 && - & 12.24 & 0.052 \\

&CC
~\cite{ranjan2019competitive} 
&& 29.00 & \second{6.92} & \second{0.018} && \second{13.77} & 7.97 & 0.031 \\

&DeepMatchVO
~\cite{shen2019icra} 
&& \second{27.08} & 9.91 & 0.038 && 24.44 & 12.18 & 0.059 \\

&PoseGraph
~\cite{li2019pose} 
&& - & 8.10 & 0.028 && - & 12.90 & 0.032 \\

&MonoDepth2
~\cite{godard2019digging} 
&& 55.47 & 11.47 & 0.032 && 20.46 & \second{7.73} & 0.034 \\

&SC-SfMLearer~\cite{bian2019unsupervised}
&& - & 11.2 & 0.034 && - & 10.1 & 0.050 \\

&Ours
&& \best{11.30} & \best{3.49} & \best{0.010} && \best{11.80} & \best{5.81} & \best{0.018} \\

\bottomrule
\end{tabular}

}

\label{tab:kitti_odom}
\end{table*}

\para{Comparison with the state-of-the-art methods.}
For comparison, we select several state-of-the-art methods, including the monocular version of ORB-SLAM2~\cite{murORB2} 
(denoted as ORB-SLAM2-M) 
(with and without loop closure optimization), several supervised learning methods~\cite{teed2018deepv2d,wang2017deepvo,wang2018end,xue2018guided,xue2019beyond}, and some other self-supervised methods~\cite{bian2019unsupervised,godard2019digging,li2018undeepvo,li2019pose,mahjourian2018unsupervised,ranjan2019competitive,shen2019icra,wang2019recurrent,yin2018geonet,zhan2018unsupervised,zhou2017unsupervised}.
Note that all supervised methods are trained on Sequence 00, 02, 08, 09 of the KITTI Odometry dataset~\cite{Geiger2012CVPR}, except DeepV2D~\cite{teed2018deepv2d}, which is trained on the Eigen split of KITTI raw dataset~\cite{Geiger2013IJRR}.
As we can see in~\tabref{kitti_odom}, our final model outperforms other self-supervised methods by a significant margin.
In particular, our method outperforms the recent proposed SC-SfMLearner~\cite{bian2019unsupervised}, which aims to reconstruct the scale-consistent camera trajectory.
This indicates that the explicit long-term modeling used in our approach is more effective than propagating the geometric constraint among overlapping short snippets in SC-SfMLearner~\cite{bian2019unsupervised}.
Our model also compares favorably with the geometric method and outperforms all supervised methods except Xue~\etal~\cite{xue2019beyond} on Sequence 10.

\para{Results on additional KITTI sequences.}
From the raw data of the KITTI dataset, we select 18 short sequences that have no overlaps with either the training or test split of the KITTI Odometry dataset.\footnote{The sequence names are available in the supplementary material.}
We then apply the same pre-trained models on these test sequences.
As we can see in~\tabref{kitti_raw}, our method outperforms other learning-based methods and even compares favorably with ORB-SLAM2-M methods in terms of RMSE and relative translation error.

\begin{table}[htbp]
\centering
\caption{\tb{Average results on 18 additional KITTI sequences.}
The results of ORB-SLAM2-M methods are the medians of 5 times.
The best performance of each block is in \best{bold}, and the second best is \second{underlined}.
}

\resizebox{0.75\textwidth}{!}{

\begin{tabular}{llccc}
\toprule
\multicolumn{2}{c}{Method}
& RMSE (m) & Rel. trans (\%) & Rel. rot. (deg/m) 
\\

\midrule
\multirow{2}{*}{Geo.}
&ORB-SLAM2-M (w/o LC)
~\cite{murORB2} 
& \best{7.17} & \best{9.41} & \best{0.008} \\

&ORB-SLAM2-M
~\cite{murORB2} 
& \second{8.12} & \second{10.64} & \best{0.008} \\

\midrule
Sup.
&DeepV2D
~\cite{teed2018deepv2d}
& 10.94 & 11.81 & 0.028 \\

\midrule
\multirow{6}{*}{Self-Sup.}
&SfMLearner
~\cite{zhou2017unsupervised} 
& 10.79 & 13.82 & 0.041
\\

&GeoNet
~\cite{yin2018geonet} 
& 14.41 & 18.99 & 0.076
\\

&CC
~\cite{ranjan2019competitive} 
& \second{7.51} & \second{10.49} & \second{0.024}
\\

&DeepMatchVO
~\cite{shen2019icra}
& 8.53 & 12.76 & 0.033
\\

&MonoDepth2
~\cite{godard2019digging} 
& \second{7.51} & 11.99 & 0.028
\\

&Ours
& \best{6.47} & \best{9.99} & \best{0.019}
\\

\bottomrule
\end{tabular}

}

\label{tab:kitti_raw}
\end{table}

\begin{table}[!h]
\centering
\caption{\tb{Results on KITTI Odometry official test split.} 
Since ground truth trajectories are not publicly available, we use estimations from the stereo version of ORB-SLAM2 as pseudo ground truth.
The best performance of each block is in \best{bold}, and the second best is \second{underlined}.
}

\resizebox{0.75\textwidth}{!}{

\begin{tabular}{llccc}
\toprule
\multicolumn{2}{c}{Method}
& RMSE (m) & Rel. trans (\%) & Rel. rot. (deg/m) 
\\

\midrule
\multirow{2}{*}{Geo.}
&ORB-SLAM2-M (w/o LC)
~\cite{murORB2} 
& \second{81.20} & \second{19.60} & \second{0.009} \\

&ORB-SLAM2-M
~\cite{murORB2} 
& \best{44.09} & \best{12.96} & \best{0.007} \\

\midrule
Sup.
&DeepV2D
~\cite{teed2018deepv2d}
& 221.33 & 24.61 & 0.041 \\

\midrule
\multirow{6}{*}{Self-Sup.}
&SfMLearner
~\cite{zhou2017unsupervised} 
& 75.00 & 26.54 & 0.045
\\

&GeoNet
~\cite{yin2018geonet} 
& 94.98 & 29.11 & 0.062
\\

&CC
~\cite{ranjan2019competitive} 
& \best{55.44} & 16.65 & 0.032
\\

&DeepMatchVO
~\cite{shen2019icra}
& 95.79 & 17.31 & 0.038
\\

&MonoDepth2
~\cite{godard2019digging} 
& 99.36 & \second{12.28} & \second{0.031}
\\

&Ours
& \second{71.63} & \best{7.28} & \best{0.014}
\\

\bottomrule
\end{tabular}

}

\label{tab:kitti_test}
\end{table}

\begin{figure}[t]
\centering

\mpage{0.47}{\includegraphics[width=1.0\linewidth]{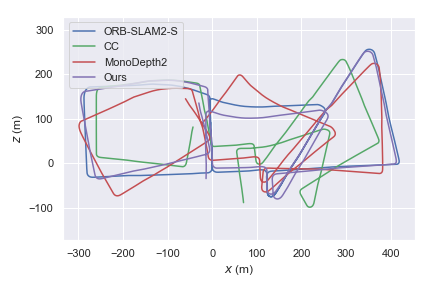}}\hfill
\mpage{0.47}{\includegraphics[width=1.0\linewidth]{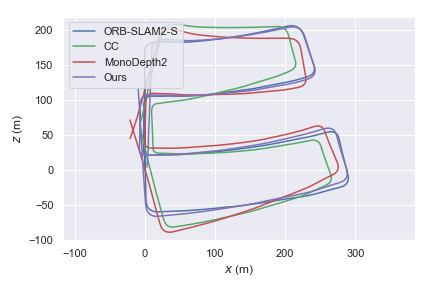}}\hfill

\vspace{-1mm}
\mpage{0.47}{(a) Seq. 13} \hfill
\mpage{0.47}{(b) Seq. 16} \hfill

\mpage{0.47}{\includegraphics[width=1.0\linewidth]{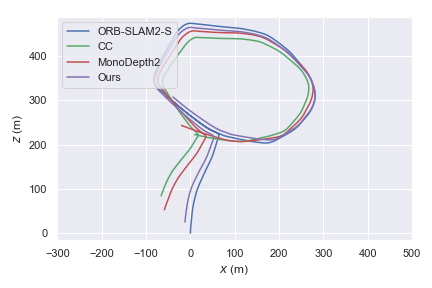}}\hfill
\mpage{0.47}{\includegraphics[width=1.0\linewidth]{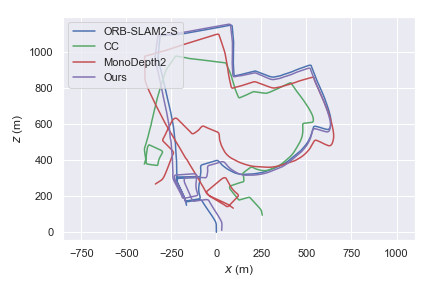}}\hfill

\vspace{-1mm}
\mpage{0.47}{(c) Seq. 18} \hfill
\mpage{0.47}{(d) Seq. 19} \hfill

\caption{\textbf{Visual comparison on the KITTI Odometry test set.}
We show the trajectories of ORB-SLAM2-S, CC~\cite{ranjan2019competitive}, MonoDepth2~\cite{godard2019digging} and our method.
Our method aligns best with the reference ORB-SLAM2-S trajectories. Best viewed in color.
}
\label{fig:sota}
\end{figure}
\para{Results on KITTI test sequences.}
Since the ground truth trajectories of Sequence 11-21 on the KITTI Odometry dataset are not available, we cannot directly recover the global scale using similarity transformation.
Thus, we choose to run the stereo version of ORB-SLAM2 (denoted as ORB-SLAM2-S), which is one of the state-of-the-art methods on these sequences.
To compare with other methods, we treat the estimations from ORB-SLAM2-S as pseudo ground truth.
As we can see in~\tabref{kitti_test},
our method achieves state-of-the-art performance among the learning-based methods and even compares favorably with ORB-SLAM2-M in terms of the relative translation error.
We visualize trajectories of four sequences in~\figref{sota}.
As we can see, our method better aligns with the reference trajectories from ORB-SLAM2-S.
We also submit the globally aligned results to the KITTI evaluation server and get similar numbers, which indicates using ORB-SLAM2-S as pseudo ground truth for evaluation is reasonable.
Please refer to the supplementary material for more details.

\begin{figure}[!h]
\centering

\mpage{0.47}{\includegraphics[width=0.8\linewidth]{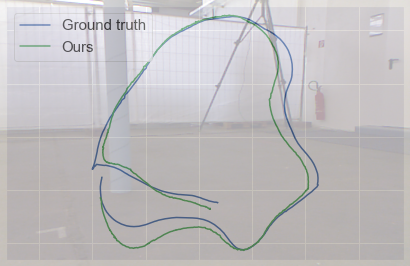}}\hfill
\mpage{0.47}{\includegraphics[width=0.8\linewidth]{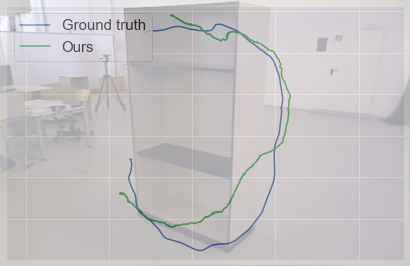}}\hfill

\vspace{-2.3mm}
\caption{\textbf{Qualitative Results on the TUM RGB-D dataset. }
We overlay the first frame of each sequence with the trajectory. Best viewed in color.
\vspace{-0.5cm}
}
\label{fig:tum}
\end{figure}
\para{Results on the TUM RGB-D dataset.}
The TUM RGB-D dataset was created to evaluate the performance of RGB-D SLAM and is thus very challenging for monocular methods.
To test our model in indoor environments, we instead compare our model with several strong baselines.
For traditional methods, we choose the monocular version of ORB-SLAM2 (denoted as ORB-SLAM2-M) and DSO~\cite{engel2017direct}.
For learning-based methods, we choose the BeyondTracking method from Xue~\etal~\cite{xue2019beyond} and the recent DeepV2D~\cite{teed2018deepv2d}.
For DeepV2D, since it is trained on another indoor dataset, in case it cannot 
generalize its scale to TUM RGB-D, we provide two options: with and without global scale alignment.

\tabref{tum_rgbd} shows that traditional methods perform well on some of the sequences, but they failed to produce results from the remaining ones due to tracking failure.
Our method outperforms the supervised baseline DeepV2D~\cite{teed2018deepv2d} on most of the sequences, but compares less favorably than the supervised VO method in~\cite{xue2019beyond}.
We conjecture that this is due to the limited amount of training data available on the TUM RGB-D dataset.
Currently, we use the same amount of training data as supervised methods. 
Adding more unlabeled video data to the training might lead to better performance for our method.
We also notice that the rolling shutter issue in this dataset makes the photometric consistency assumption less accurate, which could potentially hurt the performance of both the proposed method and DSO.

\begin{table*}[!t]
\centering
\caption{\tb{Results on the TUM-RGBD dataset~\cite{sturm2012benchmark}.}
`-' means that traditional method fails in that sequence.
The best performance of each block is in \best{bold}, and the second best is \second{underlined}.
}

\resizebox{0.85\textwidth}{!}{

\begin{tabular}{llccccccccccc}
\toprule
\multicolumn{2}{c}{Method} 
& Seq. 1 & Seq. 2 & Seq. 3 & Seq. 4 & Seq. 5 & Seq. 6 & Seq. 7 & Seq. 8 & Seq. 9 & Seq. 10 & Avg. \\

\midrule
\multirow{2}{*}{Geo.}
&ORB-SLAM2-M
~\cite{murORB2} 
& \best{0.041} & \best{0.184} & - & - & - & - & - & \best{0.057} & - & \best{0.018} & - \\

&DSO
~\cite{engel2017direct}
& - & \second{0.197} & - & \best{0.737} & - & \best{0.082} & - & \second{0.093} & \best{0.543} & \second{0.040} & - \\

\midrule

\multirow{3}{*}{Sup.}
&BeyondTracking
~\cite{xue2019beyond}
& \second{0.153} & \second{0.208} & \best{0.056} & \best{0.070} & \second{0.172} & \second{0.015} & \best{0.123} & \best{0.007} & \best{0.035} & \best{0.042} & \best{0.088} \\

&DeepV2D
~\cite{teed2018deepv2d}
& 0.232 & 0.651 & 0.186 & 0.167 & \best{0.171} & 0.029 & 0.435 & \second{0.106} & \second{0.085} & \second{0.082} & 0.214 \\

&DeepV2D (aligned)
~\cite{teed2018deepv2d}
& \best{0.087} & 0.300 & \second{0.114} & \second{0.106} & 0.181 & \best{0.013} & \second{0.380} & 0.110 & 0.094 & 0.098 & \second{0.148} \\

\midrule

Self-Sup.
&Ours & 0.192 & 0.190 & 0.083 & 0.122 & 0.177 & 0.016 & 0.219 & 0.102 & 0.179 & 0.107 & 0.139 \\

\bottomrule
\end{tabular}

}

\label{tab:tum_rgbd}
\end{table*}

\para{Discussions.}
Our learning framework is motivated by geometric VO methods. 
The FlowNet backbone mimics the {\it tracking module} to extract pair-wise image features, and the LSTMs model the {\it sequential nature} of the VO problem. 
The design of the two-layer LSTM module also resembles the {\it keyframe mechanism} of geometric VO in the sense that the second LSTM predicts the motion between a keyframe and a non-keyframe, refining the initial consecutive estimations from the first LSTM.
The cycle consistency constraint between the two-layer LSTM estimations serves as a mini {\it loop closure} to enforce the transitivity consistency of poses.
The second stage of training allows our network to explicitly optimize over \textit{long} sequences, which resembles the {\it motion-only bundle adjustment module}.
We combine the best of both geometry and learning by building a self-supervised VO framework whose components (network, loss, training scheme) are fully inspired by the well-studied geometric modules. As verified in our experiments, these geometry inspired designs lead to significantly better results than the existing self-supervised baselines.

\para{Limitations.}
Although the proposed system achieves a good camera pose estimation performance in terms of translation error, the improvement on the rotation prediction is not as substantial as the translation.
We conjecture that the large rotation error may be due to the bias within the training data. Specifically, for driving scenarios, the translational motions occur much more frequently than the rotational ones. 
Training on more diverse video sequences or synthetic data could potentially alleviate the inherent bias in the existing datasets.
Also, we observe that our system fails under the over-exposure scenarios since our method still relies on visual input to extract information.

\section{Conclusions}
\label{sec:conclusions}
\vspace{\secmargin}
In this work, we learn a monocular visual odometry system in a self-supervised manner to mimic critical modules in traditional geometric methods.
We first adopt a two-layer convolutional LSTM module to model the long-term dependency in the pose estimation.
To allow the network to see beyond \textit{short} snippets (\eg 3 or 5 frames) during the training time, we propose a stage-wise training strategy.
Combining the recurrent architecture and the proposed decoupled training scheme, our system achieves state-of-the-art performance among self-supervised methods.
In the current form, we do not have a mechanism to detect loops and perform full loop closure. In the future, we plan to study how to incorporate it into our learning framework.

\flushleft\textbf{Acknowledgment.}
This work was part of Y. Zou's internship at NEC Labs America, in San Jose.
Y. Zou and J.-B. Huang were also supported in part by NSF under Grant No. (\#1755785).

\clearpage

\bibliographystyle{splncs04}
\bibliography{references}
\clearpage
\addcontentsline{toc}{section}{Supplementary Material}
\renewcommand{\thesubsection}{\Alph{subsection}}
\section*{Supplementary Material}
\label{sec:overview}
In this supplementary document, we provide additional experimental results and information to complement the main manuscript.
First, we conduct additional ablation experiments to further validate our design choices.
Second, we show our results on the KITTI Odometry leaderboard.
Third, we show results on the KITTI Odometry training split.
Fourth, we show results on the snippet-level pose and single-view depth estimation for completeness.
Lastly, we provide the list of sequences we selected from KITTI raw data.
We also provide a demo video showing the trajectories of several challenging sequences in the KITTI Odometry dataset.
Please refer to the attached file \textbf{supp\_video.mp4}.

\subsection{Ablation Study}
In~\tabref{ablation_component}, we conduct an ablation study to validate the effectiveness of the incorporated cycle consistency constraint, pose features (from $I_0$ and $I_t$), depth features, and the memory buffer in our two-layer ConvLSTM module.
As we can see, all the components help improve performance in the first stage of training.
\begin{table*}[htbp]
\centering
\caption{\tb{Ablation study on different components} of the second-layer ConvLSTM.
The best performance is in \best{bold} and the second best is \second{underlined}.
}

\resizebox{1\textwidth}{!}{
\scriptsize
\begin{tabular}{lcccccccccccc}
\toprule
&& \multicolumn{3}{c}{Seq. 09} && 
\multicolumn{3}{c}{Seq. 10}
\\
\cmidrule{3-5} \cmidrule{7-9}
Method && RMSE (m) & Rel. trans. (\%) & Rel. rot. (deg/m) && RMSE (m) & Rel. trans. (\%) & Rel. rot. (deg/m) \\

\midrule
Two-layer ConvLSTM (w/o cycle consistency)
&& 20.37 & 5.02 & 0.016 && 16.63 & 6.88 & 0.035
\\

Two-layer ConvLSTM (w/o pose features)
&& 14.26 & 5.64 & 0.018 && 14.47 & 7.52 & 0.030
\\

Two-layer ConvLSTM (w/o depth features)
&& \second{11.53} & \second{4.54} & 0.015 && 14.07 & \second{6.54} & 0.031 \\

Two-layer ConvLSTM (w/o memory buffer)
&& 12.54 & 5.12 & \second{0.014} && \second{13.96} & 7.20 & \second{0.026} \\

Two-layer ConvLSTM
&& \best{9.77} & \best{4.23} & \best{0.013} && \best{12.68} & \best{6.02} & \best{0.023} \\

\bottomrule
\end{tabular}

}

\label{tab:ablation_component}
\end{table*}

In~\tabref{ablation_length}, we conduct an ablation study to show the performance of different input sequence lengths of the second stage of training.
Our results show that the performance gradually improves as we increase the number of input frames during training.
When the number of frames reaches the GPU memory limitations (\eg our default setting, 97-frame), we achieve the best performance.
Training the model on a GPU with larger memory could potentially improve the performance further.
\begin{table*}[htbp]
\centering
\caption{\tb{Ablation study on different input sequence length} of the second-stage of training.
The best performance is in \best{bold} and the second best is \second{underlined}.
}

\resizebox{1\textwidth}{!}{
\scriptsize

\begin{tabular}{lcccccccccccc}
\toprule
&& \multicolumn{3}{c}{Seq. 09} && 
\multicolumn{3}{c}{Seq. 10}
\\
\cmidrule{3-5} \cmidrule{7-9}
Method && RMSE (m) & Rel. trans. (\%) & Rel. rot. (deg/m) && RMSE (m) & Rel. trans. (\%) & Rel. rot. (deg/m) \\

\midrule
49-frame
&& 12.50 & 3.83 & \second{0.011} && 12.30 & 5.99 & \best{0.018} \\

73-frame
&& \second{12.42} & \second{3.69} & \best{0.010} && \second{12.06} & \second{5.89} & \best{0.018} \\

97-frame (default)
&& \best{11.30} & \best{3.49} & \best{0.010} && \best{11.80} & \best{5.81} & \best{0.018} \\

\bottomrule
\end{tabular}

}

\label{tab:ablation_length}
\end{table*}

\subsection{Results on KITTI Odometry Test Set}
In~\tabref{kitti_server}, we provide results on the KITTI Odometry leaderboard.
It may be observed that the performance of our method is close to Table 5 in the main manuscript.
This suggests that using ORB-SLAM2-S as pseudo ground truth is a reasonable choice for evaluation.

In addition to our method, we select two state-of-the-art self-supervised methods (CC~\cite{ranjan2019competitive} and MonoDepth2~\cite{godard2019digging}) and submit the estimated results to the server as well.
Our method compares favorably with these two self-supervised methods.
Our method also outperforms the supervised method DeepVO~\cite{wang2017deepvo} by a large margin.
\begin{table}[htbp]
\centering
\caption{\tb{Results on KITTI Odometry leaderboard.} 
Note that we use the estimations from ORB-SLAM2-S~\cite{murORB2} to align scale globally for the self-supervised methods.
}

\begin{tabular}{llcc}
\toprule
&Method 
& Rel. trans (\%) & Rel. rot. (deg/m) 
\\

\midrule
\parbox[t]{20mm}{\multirow{3}{*}{{Geo.}}}
&ORB-SLAM2-S~\cite{murORB2} 
& 1.70 & 0.0028 \\

&VISO2-M~\cite{geiger2011stereoscan}
& 11.94 & 0.0234
\\

&VISO2-M+GP~\cite{geiger2011stereoscan,song2014robust}
& 7.46 & 0.0245
\\

\hline
\parbox[t]{20mm}{\multirow{1}{*}{{Sup.}}}
&DeepVO~\cite{wang2017deepvo}
& 24.55 & 0.0489
\\

\hline
\parbox[t]{20mm}{\multirow{3}{*}{{Self-Sup.}}}
&CC~\cite{ranjan2019competitive}
& 16.06 & 0.0320
\\

&MonoDepth2~\cite{godard2019digging}
& 12.59 & 0.0312 
\\

&Ours
& 7.40 & 0.0142
\\

\bottomrule
\end{tabular}

\label{tab:kitti_server}
\end{table}

\begin{figure}[t]
\centering

\mpage{0.47}{\includegraphics[width=1.0\linewidth]{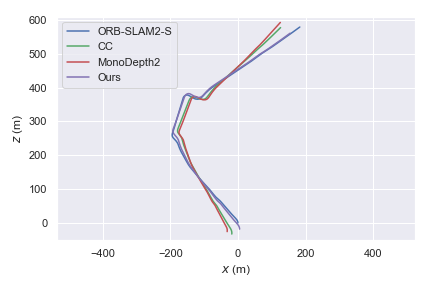}}\hfill
\mpage{0.47}{\includegraphics[width=1.0\linewidth]{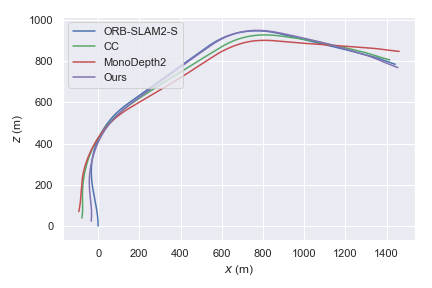}}\hfill

\mpage{0.47}{(a) Seq. 11} \hfill
\mpage{0.47}{(b) Seq. 12} \hfill

\mpage{0.47}{\includegraphics[width=1.0\linewidth]{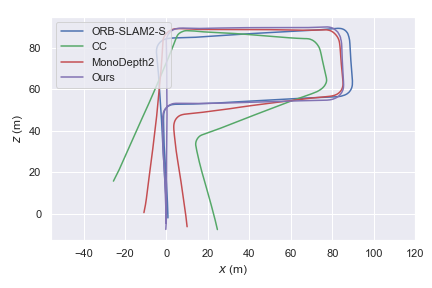}}\hfill
\mpage{0.47}{\includegraphics[width=1.0\linewidth]{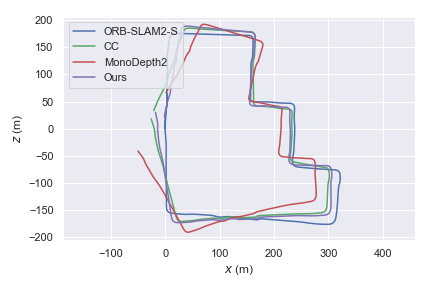}}\hfill

\mpage{0.47}{(a) Seq. 14} \hfill
\mpage{0.47}{(b) Seq. 15} \hfill

\mpage{0.47}{\includegraphics[width=1.0\linewidth]{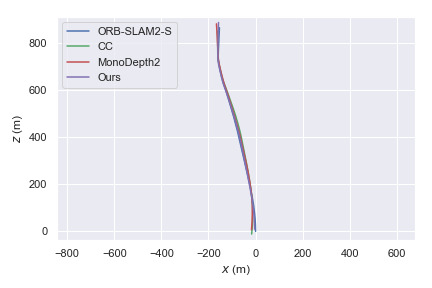}}\hfill
\mpage{0.47}{\includegraphics[width=1.0\linewidth]{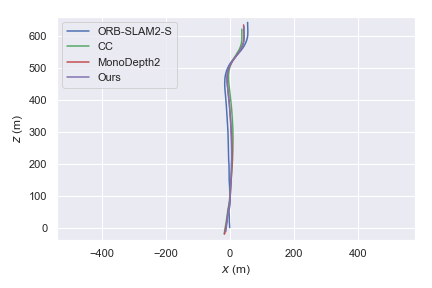}}\hfill

\mpage{0.47}{(a) Seq. 17} \hfill
\mpage{0.47}{(b) Seq. 20} \hfill

\mpage{0.47}{\includegraphics[width=1.0\linewidth]{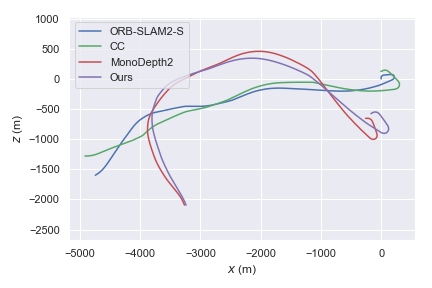}}\hfill

\mpage{0.47}{(b) Seq. 21} \hfill

\caption{\textbf{Visual comparison on the KITTI Odometry test set.}
We show the trajectories of ORB-SLAM2-S, CC~\cite{ranjan2019competitive}, MonoDepth2~\cite{godard2019digging} and our method.
Our method aligns best with the reference ORB-SLAM2-S trajectories.
}
\label{fig:sota_supp}
\end{figure}

In~\figref{sota_supp}, we show qualitative results on the remaining 7 sequences (other than those shown in the main manuscript) from the KITTI Odometry test set.
Our method aligns best with the reference ORB-SLAM2-S trajectories.

\subsection{Results on KITTI Odometry Training Set}
\begin{table}[htbp]
\centering

\caption{\textbf{Pose evalution} on \textit{training split} of KITTI Odometry dataest~\cite{Geiger2012CVPR}.
The results of ORB-SLAM2-M methods are the medians of 5 times.
`-' means the results are not available from that paper.
For DeepV2D~\cite{teed2018deepv2d}, SfMLearner~\cite{zhou2017unsupervised}, GeoNet~\cite{yin2018geonet}, CC~\cite{ranjan2019competitive}, DeepMatchVO~\cite{shen2019icra}, and MonoDepth2~\cite{godard2019digging}, we take the pre-trained models and run on the sequences to get the results.
The best performance of each block is in \best{bold}, and the second best is \second{underlined}.
}

\resizebox{\textwidth}{!}{
\begin{tabular}{llccccccccc}
\toprule
&\textbf{RMSE (m)} & Seq. 00 & Seq. 01 & Seq. 02 & Seq. 03 & Seq. 04 & Seq. 05 & Seq. 06 & Seq. 07 & Seq. 08 \\
\midrule
\parbox[t]{20mm}{\multirow{2}{*}{{Geo.}}}
&ORB-SLAM2-M (w/o LC) & \second{54.94} & \second{568.63} & \second{58.55} & \best{1.41} & \second{2.41} & \second{29.32} & \second{51.87} & \second{16.83} & \best{36.90} \\
&ORB-SLAM2-M & \best{9.02} & \best{529.28} & \best{17.96} & \second{2.07} & \best{1.56} & \best{5.20} & \best{14.07} & \best{2.88} & \second{37.83} \\
\midrule
Sup.
&DeepV2D~\cite{teed2018deepv2d} & 101.08 & 484.87 & 121.02 & 3.62 & 8.86 & 35.23 & 113.31 & 12.86 & 55.69 \\
\midrule
\parbox[t]{20mm}{\multirow{6}{*}{{Self-Sup.}}}
&SfMLearner~\cite{zhou2017unsupervised} & 97.81 & 108.09 & 152.15 & 7.47 & 2.49 & 48.13 & 39.56 & 21.28 & 32.56 \\

&GeoNet~\cite{yin2018geonet} & 148.81 & 168.90 & 293.46 & 17.58 & 7.26 & 86.94 & 17.69 & 13.88 & 138.00 \\

&CC~\cite{ranjan2019competitive} & 68.31 & 50.41 & \second{59.19} & 8.89 & 2.25 & 22.49 & 13.02 & 11.31 & 49.29 \\

&DeepMatchVO~\cite{shen2019icra} & \second{51.34} & 85.96 & 127.99 & 11.03 & 3.09 & 27.59 & 20.98 & 16.71 & \second{38.71} \\

&MonoDepth2~\cite{godard2019digging} & 82.05 & \best{30.81} & 86.64 & \second{2.40} & \best{2.00} & \second{21.49} & \best{5.16} & \second{10.42} & 51.83 \\

&Ours & \best{13.13} & \second{41.38} & \best{12.61} & \best{1.61} & \second{2.22} & \best{8.24} & \second{9.16} & \best{9.92} & \best{13.98} \\

\bottomrule
\end{tabular}
}

\vspace{2mm}
\resizebox{\textwidth}{!}{
\begin{tabular}{llccccccccc}
\toprule
&\textbf{Rel. trans (\%)} & Seq. 00 & Seq. 01 & Seq. 02 & Seq. 03 & Seq. 04 & Seq. 05 & Seq. 06 & Seq. 07 & Seq. 08 \\
\midrule
\parbox[t]{20mm}{\multirow{2}{*}{{Geo.}}}
&ORB-SLAM2-M (w/o LC) & \second{14.11} & \second{131.75} & \second{12.70} & \best{1.21} & \second{2.40} & \second{9.12} & \second{18.50} & \second{10.34} & \best{9.72} \\
&ORB-SLAM2-M & \best{3.23} & \best{125.63} & \best{3.69} & \second{1.73} & \best{1.97} & \best{2.31} & \best{5.92} & \best{2.15} & \second{11.68} \\

\midrule
\parbox[t]{20mm}{\multirow{6}{*}{{Sup.}}}
&DeepVO~\cite{wang2017deepvo} & - & - & - & 8.49 & 7.19 & \second{2.62} & \second{5.42} & 3.91 & - \\

&ESP-VO~\cite{wang2018end} & - & - & - & 6.72 & 6.33 & 3.35 & 7.24 & 3.52 & - \\

&GFS-VO~\cite{xue2018guided} & - & - & - & 5.44 & \best{2.91} & 3.27 & 8.50 & \second{3.37} & - \\

&GFS-VO-RNN~\cite{xue2018guided} & - & - & - & 6.36 & 5.95 & 5.85 & 14.58 & 5.88 & - \\

&BeyondTracking~\cite{xue2019beyond} & - & - & - & \best{3.32} & \second{2.96} & \best{2.59} & \best{4.93} & \best{3.07} & - \\

&DeepV2D~\cite{teed2018deepv2d} & \best{12.38} & \best{56.26} & \best{7.79} & \second{4.07} & 8.22 & 6.35 & 16.67 & 4.96 & \best{6.63} \\

\midrule
\parbox[t]{20mm}{\multirow{6}{*}{{Self-Sup.}}}
&SfMLearner~\cite{zhou2017unsupervised} & 19.27 & 21.71 & 18.99 & 9.73 & 3.17 & 10.02 & 11.00 & 11.68 & 8.67 \\

&GeoNet~\cite{yin2018geonet} & 33.63 & 22.96 & 54.00 & 19.41 & 10.81 & 22.68 & 9.90 & 9.82 & 22.26 \\

&CC~\cite{ranjan2019competitive} & 10.42 & 15.64 & \second{8.08} & 8.49 & \second{2.90} & 5.70 & 4.38 & \second{5.91} & \second{7.16} \\

&DeepMatchVO~\cite{shen2019icra} & \second{5.31} & 29.57 & 15.94 & 9.67 & 4.15 & 7.42 & 5.69 & 7.62 & 9.43 \\

&MonoDepth2~\cite{godard2019digging} & 7.64 & \best{10.06} & 8.34 & \second{5.30} & 3.20 & \second{4.66} & \best{2.48} & \best{4.58} & 7.32 \\

&Ours & \best{2.60} & \second{13.27} & \best{2.49} & \best{1.59} & \best{2.52} & \best{2.63} & \second{2.64} & 6.43 & \best{3.61} \\

\bottomrule
\end{tabular}
}

\vspace{2mm}
\resizebox{\textwidth}{!}{
\begin{tabular}{llccccccccc}
\toprule
&\textbf{Rel. rot (deg/m)} & Seq. 00 & Seq. 01 & Seq. 02 & Seq. 03 & Seq. 04 & Seq. 05 & Seq. 06 & Seq. 07 & Seq. 08 \\
\midrule
\parbox[t]{20mm}{\multirow{2}{*}{{Geo.}}}
&ORB-SLAM2-M (w/o LC) & \best{0.003} & \best{0.010} & \best{0.003} & \best{0.002} & \best{0.002} & \best{0.002} & \second{0.003} & \best{0.003} & \best{0.003} \\
&ORB-SLAM2-M & \best{0.003} & \second{0.012} & \second{0.004} & \best{0.002} & \best{0.002} & \second{0.003} & \best{0.002} & \second{0.005} & \best{0.003} \\

\midrule
\parbox[t]{20mm}{\multirow{6}{*}{{Sup.}}}
&DeepVO~\cite{wang2017deepvo}
& - & - & - & 0.069 & 0.070 & 0.036 & 0.058 & 0.046 & - 
\\

&ESP-VO~\cite{wang2018end}
& - & - & - & 0.065 & 0.061 & 0.049 & 0.073 & 0.050 & -
\\

&GFS-VO~\cite{xue2018guided}
& - & - & - & \second{0.033} & \best{0.013} & \second{0.016} & \second{0.027} & \second{0.022} & - \\

&GFS-VO-RNN~\cite{xue2018guided}
& - & - & - & 0.036 & 0.024 & 0.025 & 0.050 & 0.026 & - \\

&BeyondTracking~\cite{xue2019beyond}
& - & - & - & \best{0.021} & \second{0.018} & \best{0.012} & \best{0.019} & \best{0.018} & - \\

&DeepV2D~\cite{teed2018deepv2d} & \best{0.051} & \best{0.051} & \best{0.030} & \best{0.021} & 0.034 & 0.027 & 0.073 & 0.030 & \best{0.031} \\

\midrule
\parbox[t]{20mm}{\multirow{6}{*}{{Self-Sup.}}}
&SfMLearner~\cite{zhou2017unsupervised} & 0.057 & 0.026 & 0.033 & 0.035 & 0.033 & 0.036 & 0.038 & 0.059 & 0.026 \\

&GeoNet~\cite{yin2018geonet} & 0.057 & 0.041 & 0.061 & 0.098 & 0.070 & 0.077 & 0.043 & 0.059 & 0.078 \\

&CC~\cite{ranjan2019competitive} & 0.035 & 0.011 & 0.016 & 0.041 & 0.012 & 0.022 & 0.008 & 0.031 & 0.023 \\

&DeepMatchVO~\cite{shen2019icra} & \second{0.013} & 0.013 & 0.024 & 0.046 & 0.020 & \second{0.017} & 0.022 & 0.037 & \second{0.012} \\

&MonoDepth2~\cite{godard2019digging} & 0.021 & \second{0.010} & \second{0.015} & \second{0.014} & \second{0.008} & \second{0.017} & \best{0.004} & \second{0.026} & 0.024 \\

&Ours & \best{0.005} & \best{0.003} & \best{0.003} & \best{0.006} & \best{0.005} & \best{0.005} & \second{0.007} & \best{0.021} & \best{0.003} \\

\bottomrule
\end{tabular}
}

\label{tab:kitti_train}
\end{table}

In~\tabref{kitti_train}, we compare the results on the training set of the KITTI Odometry dataset.
Note that all supervised methods are trained on Sequence 00, 02, 08, 09 of the KITTI Odometry dataset~\cite{Geiger2012CVPR}, except DeepV2D~\cite{teed2018deepv2d}, which is trained on the Eigen split of KITTI raw dataset~\cite{Geiger2013IJRR}.
Comparing to other self-supervised approaches, our method achieves smaller errors on the training set, indicating that the proposed system can effectively learn to model the camera pose trajectory during training time.
Our method also compares favorably against the geometric-based method ORB-SLAM2.

\begin{figure}[t]
\centering

\mpage{0.32}{\includegraphics[width=1.0\linewidth]{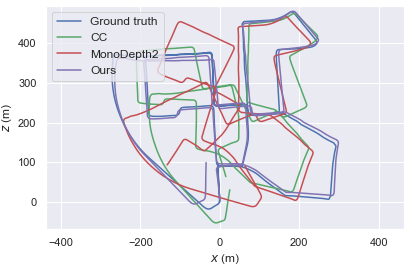}}\hfill
\mpage{0.32}{\includegraphics[width=1.0\linewidth]{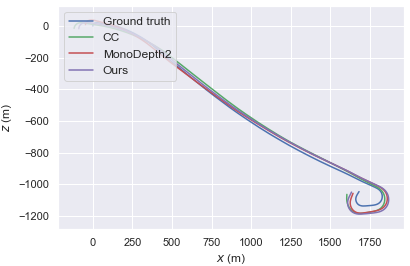}}\hfill
\mpage{0.32}{\includegraphics[width=1.0\linewidth]{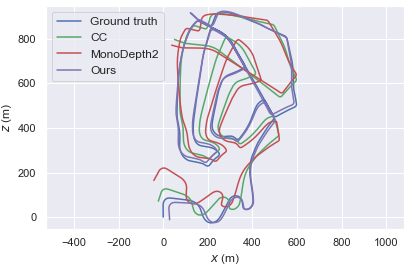}}\hfill

\mpage{0.32}{(a) Seq. 00} \hfill
\mpage{0.32}{(b) Seq. 01} \hfill
\mpage{0.32}{(b) Seq. 02} \hfill

\mpage{0.32}{\includegraphics[width=1.0\linewidth]{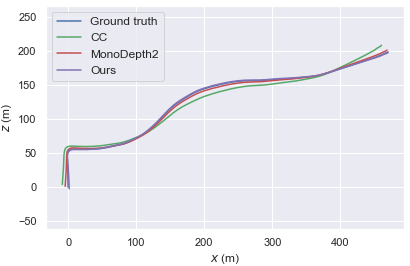}}\hfill
\mpage{0.32}{\includegraphics[width=1.0\linewidth]{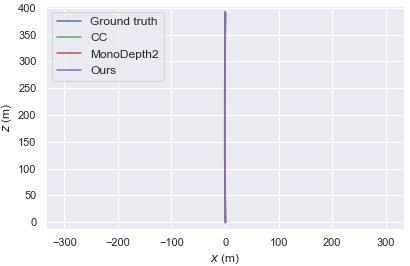}}\hfill
\mpage{0.32}{\includegraphics[width=1.0\linewidth]{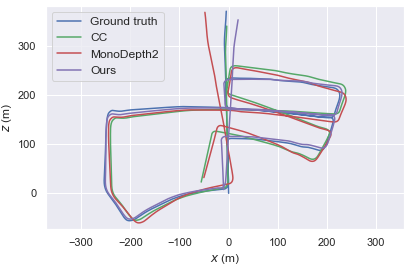}}\hfill

\mpage{0.32}{(a) Seq. 03} \hfill
\mpage{0.32}{(b) Seq. 04} \hfill
\mpage{0.32}{(b) Seq. 05} \hfill

\mpage{0.32}{\includegraphics[width=1.0\linewidth]{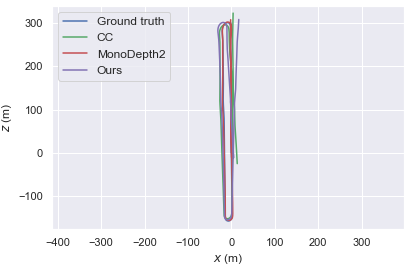}}\hfill
\mpage{0.32}{\includegraphics[width=1.0\linewidth]{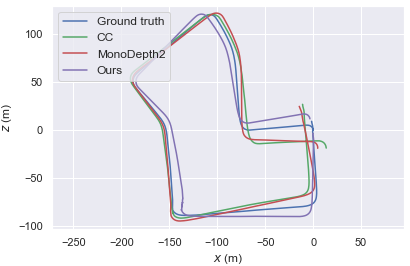}}\hfill
\mpage{0.32}{\includegraphics[width=1.0\linewidth]{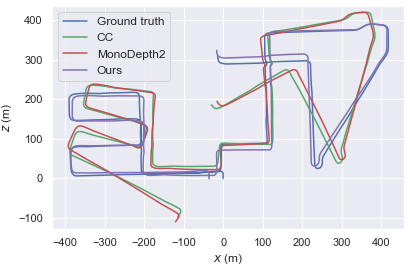}}\hfill

\mpage{0.32}{(a) Seq. 06} \hfill
\mpage{0.32}{(b) Seq. 07} \hfill
\mpage{0.32}{(b) Seq. 08} \hfill

\caption{\textbf{Visual comparison on the KITTI Odometry training set.}
We show the trajectories of ORB-SLAM2-M, CC~\cite{ranjan2019competitive}, MonoDepth2~\cite{godard2019digging} and our method.
}
\label{fig:kitti_train}
\end{figure}

In~\figref{kitti_train}, we show the qualitative results of our method on Seq. 00-08 on the KITTI Odometry dataset.

\subsection{Snippet-level Pose Results and Depth Results}
For completeness, we provide the pose estimation results when evaluating on 5-frame snippets in~\tabref{kitti_pose} and the single-view depth estimation results in~\tabref{kitti_depth}.
Note that the depth network is fixed during the second stage of training, so for the depth evaluation, we only train our model for the first stage on the Eigen split of the KITTI raw dataset.
As we can see in~\tabref{kitti_pose}, although CC~\cite{ranjan2019competitive} and DeepMatchVO~\cite{shen2019icra} achieve good results on the snippet-level, their results on the video-level are no longer the state-of-the-art.
This indicates that evaluating camera pose estimation performance on the snippet-level could be inaccurate, and thus we need to evaluate the whole trajectory to reflect the holistic performance.
In~\tabref{kitti_depth}, we also observe that our method slightly outperforms the current self-supervised state-of-the-art MonoDepth2~\cite{godard2019digging}, which indicates that a better pose estimation module could lead to a better depth estimation performance.

\begin{table}[htbp]
\centering

\caption{\textbf{5-frame snippet-level results} on KITTI Odometry datest~\cite{Geiger2012CVPR}.}

\begin{tabular}{lccc}
\toprule
 & Seq. 09 && Seq. 10\\
\midrule
ORB-SLAM (full) & 0.014$\pm$0.008 && 0.012$\pm$0.011 \\
\midrule
SfMLearner~\cite{zhou2017unsupervised} & 0.021$\pm$0.017 && 0.020$\pm$0.015 \\
vid2depth~\cite{mahjourian2018unsupervised} & 0.013$\pm$0.010 && 0.012$\pm$0.011 \\
GeoNet~\cite{yin2018geonet} & 0.012$\pm$0.007 && 
0.012$\pm$0.009 \\
DF-Net~\cite{zou2018dfnet} & 0.017$\pm$0.007 && 0.015$\pm$0.009 \\
CC~\cite{ranjan2019competitive} & 0.012$\pm$0.007 && 0.012$\pm$0.008 \\
DeepMatchVO~\cite{shen2019icra} & 0.009$\pm$0.005 && 0.008$\pm$0.007 \\
MonoDepth2~\cite{godard2019digging} & 0.017$\pm$0.008 && 0.015$\pm$0.010 \\

Ours & 0.015$\pm$0.006 && 0.015$\pm$0.009 \\

\bottomrule
\end{tabular}
\label{tab:kitti_pose}
\end{table}

\begin{table*}[htbp]
\centering
\caption{\tb{Single-view depth estimation results} on \textit{Eigen test split} of KITTI raw dataset~\cite{Geiger2013IJRR}.
}
\begin{tabular}{lccccccccc}

\toprule
& \multicolumn{4}{c}{Error metric $\downarrow$} && \multicolumn{3}{c}{Accuracy metric $\uparrow$} &
\\
\cmidrule{2-5} \cmidrule{7-9} 
Method & Abs Rel  & Sq Rel  & RMSE  & log RMSE  && $\delta < 1.25$ & $\delta < 1.25^2$ & $\delta < 1.25^3$\\

\midrule
SfMLearner~\cite{zhou2017unsupervised} & 0.208 & 1.768 & 6.856 & 0.283 && 0.678 & 0.885 & 0.957 \\
vid2depth~\cite{mahjourian2018unsupervised} & 0.163 & 1.240 & 6.220 & 0.250 && 0.762 & 0.916 & 0.968 \\
GeoNet~\cite{yin2018geonet} & 0.155 & 1.296 & 5.857 & 0.233 && 0.793 & 0.931 & 0.973 \\
DF-Net~\cite{zou2018dfnet} & 0.150 & 1.124 & 5.507 & 0.223 && 0.806 & 0.933 & 0.973 \\

CC~\cite{ranjan2019competitive} & 0.140 & 1.070 & 5.326 & 0.217 && 0.826 & 0.941 & 0.975 \\

DeepMatchVO~\cite{shen2019icra} & 0.156 & 1.309 & 5.73 & 0.236 && 0.797 & 0.929 & 0.969 \\

MonoDepth2~\cite{godard2019digging} & 0.115 & 0.903 & 4.863 & 0.193 && 0.877 & 0.959 & 0.981 \\

SC-SfMLearner~\cite{bian2019unsupervised} & 0.137 & 1.089 & 5.439 & 0.217 && 0.830 & 0.942 & 0.975 \\

Ours & 0.115 & 0.871 & 4.778 & 0.191 && 0.874 & 0.961 & 0.982 \\

\bottomrule
\end{tabular}
\label{tab:kitti_depth}
\end{table*}

\subsection{Additional KITTI Sequences}
As mentioned in the main manuscript, we selected 18 sequences from KITTI raw data to further evaluate the methods, which have no overlaps with either KITTI Odometry split or Eigen split.
We list the sequence names in~\tabref{kitti_names}.
\begin{table}[htbp]
\centering

\caption{\textbf{Names of 18 additional KITTI sequences}.}

\begin{tabular}{c}
\toprule

Sequence names \\

\midrule
2011\_09\_26\_drive\_0036 \\

2011\_09\_26\_drive\_0086 \\

2011\_09\_26\_drive\_0101 \\

2011\_09\_26\_drive\_0117 \\

2011\_09\_29\_drive\_0071 \\

2011\_10\_03\_drive\_0047 \\

2011\_09\_26\_drive\_0059 \\

2011\_09\_26\_drive\_0027 \\

2011\_09\_26\_drive\_0009 \\

2011\_09\_26\_drive\_0013 \\

2011\_09\_26\_drive\_0029 \\

2011\_09\_26\_drive\_0064 \\

2011\_09\_26\_drive\_0084 \\

2011\_09\_26\_drive\_0096 \\

2011\_09\_26\_drive\_0106 \\

2011\_09\_26\_drive\_0056 \\

2011\_09\_26\_drive\_0023 \\

2011\_09\_26\_drive\_0093 \\

\bottomrule
\end{tabular}
\label{tab:kitti_names}
\end{table}

\end{document}